\documentclass[11pt]{article}

\usepackage[]{acl}

\usepackage{times}
\usepackage{latexsym}
\usepackage[T1]{fontenc}
\usepackage[utf8]{inputenc}
\usepackage{microtype}
\usepackage{inconsolata}
\usepackage{graphicx}
\usepackage{booktabs}
\usepackage{multirow}
\usepackage{amsmath}
\usepackage{amssymb}
\usepackage{authblk}

\title{Cross-Lingual Token Arbitrage: Optimizing Code Agent Context Windows via Local LLM Preprocessing}

\author{Mehmet Utku Çolak}
\affil{Department of Computer Engineering, Istanbul Technical University \\ Istanbul, Turkey \\ colakme19@itu.edu.tr}

\begin{document}
\maketitle
\begin{abstract}
AI-assisted coding agents are bottlenecked by input-token cost. Two pathologies of raw human input drive most of the overhead: \textit{tokenization overhead} ($2$--$3\times$ more tokens for non-English text) and \textit{structural entropy} in conversational prompts (greedy retrieval and verbose decoding). Existing methods act reactively, compressing already-bloated contexts or cascading on failure.

We introduce a pre-flight, edge-side prompt-rewriting middleware between the developer and the cloud agent. A local \texttt{Llama 3.2} (3B) via \texttt{Ollama} performs three operations in a sub-second pass: cross-lingual translation into the cheaper English token space, structural rewriting into a compact \texttt{[CONTEXT]}/\texttt{[TASK]} format, and a regex-validated rewrite-with-fallback guarded by a $5\%$ token-budget threshold so the optimized payload is never larger than the original.

We evaluate on \textit{OMH-Polyglot}, the polyglot-specification arm of our \textit{OMH} (\textbf{O}ck-\textbf{M}ultilingual-\textbf{H}ard) benchmark family ($200$ instances translated into Turkish, Arabic, Chinese, or code-switched mixtures; per-prompt tokenization-overhead ratio $2.05\times$ on average, max $6.15\times$). Across three commercial backends with three runs per condition, the middleware reduces prompt tokens by $34$--$47\%$ and total tokens by $8.3\%$, $8.3\%$, and $18.8\%$ on \texttt{gpt-3.5-turbo}, \texttt{gpt-4o}, and \texttt{gemini-2.5-flash-lite}. Accuracy is $99.5\%$, $99.5\%$ ($+1.17$\,pp), and $98.0\%$ ($+3.00$\,pp). A \textit{names\_only} ablation isolates the function-name oracle: alone it decreases accuracy by $18$--$20$\,pp on the two weaker backends and is neutral on \texttt{gpt-4o}, attributing the Gemini gain to the rewriter. Against LLMLingua-2 \cite{pan2024llmlingua2} at matched rate, our middleware strictly Pareto-dominates on OckScore across all three backends. Dollar-cost outcomes are asymmetric: $-12.4\%$ on Gemini, $-0.4\%$ on \texttt{gpt-4o}, and $+15.1\%$ on \texttt{gpt-3.5-turbo}.
\end{abstract}

\section{Introduction}
AI-assisted IDEs such as Cursor, GitHub Copilot, and Claude Code now operate over repositories that exceed the model's context window, so each chat turn inflates the prompt with retrieved code chunks, prior turns, and unstructured user instructions. At deployment scale, per-query overhead compounds into substantial financial and environmental cost \cite{du2026ockbench, peng2025coffe}.

A large fraction of this cost is paid \textit{before} the cloud model runs, and originates from two compounding pathologies of the raw input. (i) \textbf{Tokenization overhead}: subword tokenizers used by frontier LLMs (e.g., \texttt{cl100k\_base}) are statistically optimized for English; equivalent content in Turkish, Chinese, or Arabic can require $2$--$3\times$ more tokens \cite{petrov2023language, teklehaymanot2025tokenization}. (ii) \textbf{Structural entropy}: developer prompts are conversational and vague, driving Retrieval-Augmented Generation (RAG) into greedy retrieval \cite{jimenez2024swebench} and inducing verbose decoding \cite{ahuja2025efficientxlang}.

Existing remedies act reactively. Post-hoc token compressors \cite{jiang-etal-2023-llmlingua, shi2025longcodezip} prune an already-bloated context, paying perplexity-computation overhead inside the interactive loop. Cost-aware prompt-search frameworks \cite{taherkhani2025epic, zehle2025capo} are offline and require expensive cloud-side iterations on the API they aim to cheapen. Local--cloud cascading \cite{lu2026mccom, chen2024modelcascading} falls back to the large model with the original, unoptimized prompt the moment the small model fails. None of these proactively neutralize tokenization overhead, and none enforce structural discipline \textit{before} retrieval.

We propose a \textbf{pre-flight, edge-side prompt rewriter} placed between the developer and the cloud agent. A local \texttt{Llama 3.2} (3B) \cite{meta2024llama32} running through \texttt{Ollama} intercepts every raw query and, in a single sub-second pass, performs: (i) \textit{Cross-Lingual Token Arbitrage}---translating non-English text into the cheaper English token space; (ii) \textit{Structural Rewriting} into a compact label-bracketed \texttt{[CONTEXT]}/\texttt{[TASK]} format (a \textit{Bi-Block} in the benchmark proxy, optionally extended to a Tri-Block with \texttt{[CONSTRAINTS]} in the IDE extension); and (iii) a regex-validated rewrite-with-fallback routine guarded by a $5\%$ token-budget threshold, so the optimized payload is never larger than the original. The pipeline runs on local hardware and incurs zero recursive cloud-API cost.

\paragraph{Contributions.}
\begin{itemize}
    \item We formalize \textit{Cross-Lingual Token Arbitrage} as an architectural mechanism (rather than an empirical observation \cite{ahuja2025efficientxlang}) and operationalize it as the first stage of a pre-flight middleware.
    \item We introduce a label-bracketed prompt schema (Bi-Block for benchmarking; Tri-Block extension for IDE deployments) and a regex-validated rewrite-with-fallback guarded by a $5\%$ token-budget threshold, so the optimizer can never inflate the cloud-billed payload.
    \item We release the \textit{OMH} benchmark family: \textit{OMH-Wrapped} (English spec with multilingual wrappers) and \textit{OMH-Polyglot} (fully translated or code-switched specs), each $200$ instances over twenty hard cores and ten style indices. Polyglot raises the per-prompt tokenization-overhead ratio from $0.96\times$ to $2.05\times$.
    \item We provide an open implementation and report results on three commercial models, three runs each, showing prompt-token reductions of $34$--$47\%$ and total-token reductions of $8.3$--$18.8\%$, with accuracy unchanged or improved by up to $+3.00$\,pp. A \textit{names\_only} ablation isolates the function-name oracle (it decreases accuracy by $18$--$20$\,pp on the two weaker backends), attributing the accuracy result to the rewriter. A head-to-head comparison against LLMLingua-2 \cite{pan2024llmlingua2} at matched compression rate shows strict Pareto-dominance on OckScore across all three backends.
\end{itemize}

\section{Related Work}
\label{sec:related}

\paragraph{Prompt optimization and compression.} APE \cite{zhou2023large} and APO \cite{pryzant-etal-2023-automatic} search for instructions to maximize accuracy but ignore cost. Token-level compressors prune unimportant tokens: SelectiveContext \cite{li-etal-2023-compressing} via self-information, LLMLingua \cite{jiang-etal-2023-llmlingua} via a dynamic budget controller, and EHPC \cite{fei2025efficient} via attention-head skimming. AgentDiet \cite{xiao2026reducing} summarizes expired trajectory history. These either depend on perplexity from a reference LM (recursive cost) or prune already-bloated context after the fact.

\paragraph{Cost-aware search and cascading.} EPiC \cite{taherkhani2025epic} and CAPO \cite{zehle2025capo} are offline prompt-search frameworks requiring repeated cloud-side iterations. Model cascading \cite{chen2024modelcascading} and the local--cloud MCCom framework \cite{lu2026mccom} forward the raw prompt to the large model on escalation. EfficientXLang \cite{ahuja2025efficientxlang} reports that translating reasoning into English yields up to $40\%$ token reductions, but treats this as observation rather than an applied middleware.

\paragraph{Repository-level context and benchmarks.} Repository-level generation requires reasoning across files \cite{liu2024graphcoder}. RepoCoder \cite{zhang-etal-2023-repocoder} iteratively retrieves based on intermediate drafts; GraphCoder \cite{liu2024graphcoder} uses control- and data-flow graphs. SWEzze \cite{mechtaev2026compressing} reduces SWE-Bench token budgets by $>50\%$ while improving issue resolution. Vague prompts trigger greedy RAG over-fetching and dependency hallucinations \cite{lehai2026repository, blackduck2026ossra, morphllm2026context}. All act after retrieval. On the evaluation side, HumanEval, MBPP, and SWE-Bench \cite{jimenez2024swebench} score correctness but ignore cost; OckBench \cite{du2026ockbench} and COFFE \cite{peng2025coffe} jointly measure accuracy and efficiency but remain post-hoc measurement tools.

\section{Methodology}
\label{sec:methodology}

The middleware is a pre-flight layer that intervenes before the RAG cycle in agentic IDEs. It shifts prompt-refinement compute to a local Small Language Model (SLM), neutralizing linguistic and structural overhead before cloud dispatch.

\subsection{Architecture}
The system is a TypeScript gateway that intercepts every IDE query and forwards a rewritten payload to the cloud backend. The local reasoning layer runs \texttt{Llama 3.2} (3B) \cite{meta2024llama32} via an \texttt{Ollama} instance and is responsible for cross-lingual translation, removal of conversational filler, and deterministic structural enforcement. The cloud execution layer (e.g., \texttt{gpt-4o}) receives only the rewritten Bi-Block.

\subsection{Cross-Lingual Token Arbitrage}
Byte-Pair Encoding tokenizers such as \texttt{cl100k\_base} are biased toward English \cite{teklehaymanot2025tokenization}; non-English text uses more tokens for the same semantic content \cite{petrov2023language}. Table~\ref{tab:tokenization-overhead} summarizes the relative per-character cost. The middleware translates non-English inputs to English locally before dispatch, exploiting the result of \cite{ahuja2025efficientxlang} that English reasoning traces are more compact than equivalents in morphologically rich languages.

\begin{table}[h]
\centering
\small
\begin{tabular}{lc}
\toprule
\textbf{Language} & \textbf{Relative Tokenization Cost} \\
\midrule
English & 1.00 \\
Spanish & 1.50 \\
Italian & 1.60 \\
Turkish & 2.16 \\
Chinese & 2.41 \\
Bulgarian & 2.60 \\
Arabic & 3.00 \\
\bottomrule
\end{tabular}
\caption{Relative tokenization cost compared to English for \texttt{cl100k\_base}.}
\label{tab:tokenization-overhead}
\end{table}

\subsection{Structural Rewriting: Bi-Block and Tri-Block}
\label{sec:structure}
We deploy two variants of a label-bracketed schema. The empirical results in this paper use the Bi-Block.

\paragraph{Bi-Block (benchmark proxy).} The proxy emits two labeled blocks per task. \texttt{[CONTEXT]} contains a single constant sentence (``\textit{Python coding task; tests are authoritative.}'')---a generic preamble, not workspace-derived metadata. \texttt{[TASK]} contains an SLM-produced one- or two-sentence English instruction, followed by the original \texttt{assert} lines copied verbatim. The SLM is prompted to translate non-English text, strip filler, and mention every required function name verbatim; it is forbidden from emitting code or section labels. Function names are additionally extracted from the \texttt{assert} lines by a deterministic regex and supplied to the SLM as a hint.

\paragraph{Tri-Block (IDE extension only).} The IDE extension targets a setting where constraints are not given as executable tests, and emits a third \texttt{[CONSTRAINTS]} block containing style guides, naming conventions, performance targets, or security notes extracted from the raw query. Results in this paper bear only on the Bi-Block.

\paragraph{Decoding.} In both variants the SLM is decoded greedily ($\texttt{temperature}=0$, $\texttt{top\_p}=0.9$, prediction budget $200$ tokens for the proxy and $400$ for the IDE extension) with stop sequences preventing label or code leakage.

\subsection{Validation and Token-Budget Guard}
\label{sec:guard}
Each rewrite is checked by a regex validator (\texttt{validateLight}) that triggers regeneration if (i) the output is empty or shorter than $10$ characters, (ii) a Python code block is emitted, (iii) a filled \verb|<solution>...</solution>| envelope is produced, or (iv) required block markers are missing (Tri-Block). Up to two repair attempts are allowed.

A token-budget guard estimates cost ($1$ token per non-ASCII char, $0.25$ per ASCII char) and forwards the original prompt unchanged whenever the rewrite is not at least $5\%$ smaller. If repair attempts repeatedly fail validation, the proxy logs the violations and forwards the raw prompt. The guard ensures the proxy never inflates the cloud-billed payload.

\section{Experimental Setup}
\label{sec:setup}

\subsection{The OMH Benchmark Family}
\label{sec:omh-family}
We retain the OckBench-Coding \cite{du2026ockbench} harness: each row is a natural-language \texttt{problem} string followed by authoritative Python \texttt{assert}s, so functional grading and OckScore are unchanged. We replace the \textit{content} of those rows. The default OckBench-Coding split sits near accuracy ceiling on commercial models and offers little headroom for token-efficiency mechanisms.

\textbf{OMH} (\textbf{O}ck-\textbf{M}ultilingual-\textbf{H}ard) is a family of two $200$-row surfaces. Both share the \textbf{same twenty in-house algorithmic cores} (reference implementations plus fixed \texttt{assert} suites) and the \textbf{same ten style indices} per core ($20\times 10=200$). They differ only in which parts of the prompt are non-English.

\textbf{\textit{OMH-Wrapped}} (wrapper-only) keeps the technical specification in English; each style index applies one of ten hand-written multilingual or noisy wrappers (informal Turkish, Arabic, or Chinese openings; polite or chat-register English; Turkish--English mix; etc.). The \texttt{cl100k\_base} per-prompt tokenization-overhead ratio is $0.96\times$ on the mean and $1.34\times$ at p90, so OMH-Wrapped cannot stress a fully non-English body. We retain it as a tokenizer control.

\textbf{\textit{OMH-Polyglot}} replaces the specification sentences themselves with Turkish, Arabic, Simplified Chinese, or code-switched equivalents; \texttt{assert} lines remain ASCII so the grader is unchanged. The twenty cores are interview-style tasks (e.g., $0/1$ Knapsack, longest common subsequence, sliding-window maxima, topological sort, Levenshtein distance). The ten style registers are: pure Turkish; pure Arabic (Modern Standard); pure Simplified Chinese; three pairwise sentence-alternating mixes (TR$\leftrightarrow$AR, TR$\leftrightarrow$ZH, AR$\leftrightarrow$ZH); a tri-language TR/AR/ZH rotation; and three native-narrative registers with embedded English jargon. The per-prompt tokenization-overhead ratio is $\mathbf{2.05\times}$ on the mean, $4.02\times$ at p90, and $6.15\times$ in the worst case (Table~\ref{tab:tokenization-overhead-bench}).

\begin{table}[h]
\centering
\small
\begin{tabular}{lcc}
\toprule
\textbf{Benchmark} & \textbf{Mean} & \textbf{p90} \\
\midrule
\textit{OMH-Wrapped} (control) & $0.96\times$ & $1.34\times$ \\
\textbf{\textit{OMH-Polyglot}} (ours) & $\mathbf{2.05\times}$ & $\mathbf{4.02\times}$ \\
\bottomrule
\end{tabular}
\caption{Per-prompt tokenization-overhead ratio (\texttt{cl100k\_base} tokens of the multilingual prompt divided by an English-floor equivalent). All accuracy and cost results in Section~\ref{sec:results} are on \textit{OMH-Polyglot}.}
\label{tab:tokenization-overhead-bench}
\end{table}

\paragraph{Why not other benchmarks?} HumanEval, MBPP, and SWE-Bench \cite{jimenez2024swebench} score \texttt{pass@k} without measuring the cost of correctness. FLORES, XL-Sum, and other translation benchmarks measure fidelity, not downstream code outcomes under multilingual input. Default OckBench-Coding \cite{du2026ockbench} hits an accuracy ceiling.

\subsection{Baselines and Implementation}
We evaluate against three commercial cloud backends spanning the cost--capability frontier: \texttt{gpt-3.5-turbo}, \texttt{gpt-4o}, and \texttt{gemini-2.5-flash-lite}. The local middleware uses a headless OpenAI-compatible proxy running \texttt{Llama 3.2} (3B) via \texttt{Ollama}; only the upstream cloud endpoint changes. For the \textit{Raw} arm, OckBench dispatches the multilingual prompt directly to the cloud API with a fixed system prompt asking for a focused Python solution. For the \textit{Ours} arm, OckBench routes through our proxy; the system prompt is identical across both arms, so any delta is attributable to the user-message rewrite alone. We run three independent repeats per (model $\times$ pipeline) cell at $\texttt{temperature}=0$ to characterize residual API non-determinism, and report run means. All cloud calls use a $4096$-token output cap.

\subsection{Metrics}
\label{sec:metrics}
For every (model $\times$ pipeline) cell we report: (i) \textbf{Accuracy}: fraction of tasks whose code passes every \texttt{assert} ($N=200$); (ii) \textbf{Token counts}: total, prompt, and answer summed across $200$ tasks, plus percent change vs.\ Raw; (iii) \textbf{OckScore} \cite{du2026ockbench}, defined as
\begin{equation}
    \mathrm{OckScore} = \mathrm{Accuracy} - 10\cdot\log_{10}\!\Big(\tfrac{\bar{T}}{T_{\text{ref}}}\Big),
    \label{eq:ockscore}
\end{equation}
with $\mathrm{Accuracy}\in[0,100]$, $\bar{T}$ the mean tokens-per-task, and $T_{\text{ref}}$ a harness-reported reference budget; (iv) \textbf{Wall-clock} per $200$-task run.

\paragraph{Statistical scope.} We report descriptive contrasts, not significance tests. OMH-Polyglot is $20$ algorithm clusters $\times$ $10$ style transforms with shared reference solutions and \texttt{assert} suites per cluster, so the effective independent-trial count is closer to $20$ than $200$. A defensible inferential analysis requires a per-algorithm cluster bootstrap; we list this as required follow-up. The three repeats per cell were near-deterministic (Gemini deterministic; \texttt{gpt-3.5-turbo} varied by $<10^{-3}$; only one of three \texttt{gpt-4o} Raw runs differed by a single task) and act as a determinism probe rather than a sample for population inference.

\section{Results}
\label{sec:results}

Table~\ref{tab:main-results} reports run means; run-level sample standard deviations are uniformly small (accuracy $\le 1.15$\,pp; token columns $<0.5\%$ of the mean) and act as a determinism probe rather than inferential standard errors.

\begin{table*}[t]
\centering
\small
\setlength{\tabcolsep}{4pt}
\begin{tabular}{llccccc}
\toprule
\textbf{Cloud Agent} & \textbf{Pipeline} & \textbf{Acc.\ (\%)} & \textbf{Total Tokens} & \textbf{Prompt} & \textbf{Answer} & \textbf{OckScore $\uparrow$} \\
\midrule
\multirow{3}{*}{\texttt{gpt-3.5-turbo}}        & Raw         & $99.50$          & $94{,}338$                          & $53{,}713$          & $40{,}625$          & $99.04$ \\
                                               & LLMLingua-2 & $77.33$          & $85{,}026$                          & $42{,}314$          & $42{,}712$          & $76.91$ \\
                                               & Ours        & $\mathbf{99.50}$ & $\mathbf{86{,}474}$ ($-8.3\%$)      & $28{,}661$          & $57{,}812$          & $\mathbf{99.08}$ \\
\midrule
\multirow{3}{*}{\texttt{gpt-4o}}               & Raw         & $98.33$          & $139{,}085$                         & $43{,}565$          & $95{,}520$          & $97.66$ \\
                                               & LLMLingua-2 & $97.17$          & $124{,}790$                         & $37{,}281$          & $87{,}509$          & $96.56$ \\
                                               & Ours        & $\mathbf{99.50}$ & $\mathbf{127{,}594}$ ($-8.3\%$)     & $28{,}776$          & $98{,}819$          & $\mathbf{98.88}$ \\
\midrule
\multirow{3}{*}{\texttt{gemini-2.5-flash-lite}}& Raw         & $95.00$          & $116{,}653$                         & $44{,}918$          & $71{,}735$          & $94.43$ \\
                                               & LLMLingua-2 & $35.00$          & $243{,}193$                         & $37{,}932$          & $205{,}261$         & $33.85$ \\
                                               & Ours        & $\mathbf{98.00}$ & $\mathbf{94{,}725}$ ($-18.8\%$)     & $\mathbf{29{,}398}$ & $\mathbf{65{,}327}$ & $\mathbf{97.54}$ \\
\bottomrule
\end{tabular}
\caption{Raw cloud baseline, LLMLingua-2 \cite{pan2024llmlingua2} at matched compression rate ($r{=}0.6$, with executable \texttt{assert} block detached and re-attached verbatim so the grader is never corrupted by token pruning), and our middleware on OMH-Polyglot ($200$ rows; three runs per cell). The percentage in parentheses is the relative change in total tokens vs.\ Raw. OckScore via Equation~\ref{eq:ockscore}. Across all three backends \textit{Ours} strictly Pareto-dominates LLMLingua-2 on OckScore.}
\label{tab:main-results}
\end{table*}

\subsection{Prompt-Side and Answer-Side Behavior}
\label{sec:answer-side}
\paragraph{Prompt-side.} The prompt-token column is the most consistent finding: within each backend, \textit{Ours} reduces prompt tokens by $34$--$47\%$, with run-to-run variation under $20$ tokens out of nearly $30{,}000$. Each backend uses a different tokenizer (\texttt{cl100k\_base}, \texttt{o200k\_base}, Gemini's SentencePiece), so absolute counts are not cross-tokenizer commensurate; the defensible reading is that each tokenizer, applied to its own backend's prompts, separately exhibits a large reduction.

\paragraph{Answer-side.} Only \texttt{gemini-2.5-flash-lite} produces shorter completions ($-8.9\%$). OpenAI models produce \textit{longer} completions: $+3.5\%$ on \texttt{gpt-4o} and $+42.3\%$ on \texttt{gpt-3.5-turbo}. The most plausible explanation is that the structurally clean Bi-Block removes hedging cues and elicits more complete implementations. The prompt-side savings absorb the answer-side increase in every case, so net total tokens still fall by $8.3$--$18.8\%$. The middleware reliably compresses the input surface; it shapes the output surface only on certain backends.

\paragraph{Accuracy on \texttt{gpt-4o}.} The $+1.17$\,pp delta corresponds to a single task swinging in one of three Raw repeats, within the $1.15$\,pp run-level standard deviation; the \textit{names\_only} ablation (Section~\ref{sec:ablation}) confirms \texttt{gpt-4o} is essentially flat across arms. We do not claim it as a result.

\paragraph{Optimization floor: benchmark artifact.} Earlier preliminary experiments on \textit{OMH-Wrapped} suggested an ``Optimization Floor'' for \texttt{gpt-3.5-turbo}. On OMH-Polyglot, no regression appears: accuracy is exactly $99.50\%$ in every Raw and Ours run. The \textit{names\_only} ablation sharpens this: with SLM rewrite stripped but the multilingual body kept, \texttt{gpt-3.5-turbo} drops to $79.00\%$, a $20.5$\,pp collapse. The earlier ``floor'' was a symptom of forwarding multilingual prompts to a weaker backend without translation.

\subsection{Ablation: Isolating the Function-Name Oracle}
\label{sec:ablation}
The \textit{Ours} arm bundles structural rewriting with a deterministic oracle that extracts target function identifiers from the \texttt{assert} lines and places them at the top of \texttt{[TASK]}. To attribute the accuracy column to a specific mechanism, we run \textit{names\_only}: the SLM is bypassed and the regex-extracted name is prepended to the raw multilingual prompt as a one-line English oracle. Three repeats per cell, identical otherwise (Table~\ref{tab:ablation}).

\begin{table*}[t]
\centering
\small
\setlength{\tabcolsep}{4pt}
\begin{tabular}{llcccc}
\toprule
\textbf{Cloud Agent} & \textbf{Arm} & \textbf{Acc.\ (\%)} & \textbf{Total Tokens} & \textbf{Prompt} & \textbf{Answer} \\
\midrule
\multirow{3}{*}{\texttt{gpt-3.5-turbo}}        & \textit{raw}         & $99.50$           & $94{,}338$           & $53{,}713$           & $40{,}625$ \\
                                               & \textit{names\_only} & $79.00 \pm 1.32$  & $100{,}960$          & $60{,}093$           & $40{,}867$ \\
                                               & \textit{full}        & $\mathbf{99.50}$  & $\mathbf{86{,}474}$  & $\mathbf{28{,}661}$  & $57{,}812$ \\
\midrule
\multirow{3}{*}{\texttt{gpt-4o}}               & \textit{raw}         & $98.33$           & $139{,}085$          & $43{,}565$           & $95{,}520$ \\
                                               & \textit{names\_only} & $98.50 \pm 1.32$  & $131{,}336$          & $49{,}212$           & $82{,}124$ \\
                                               & \textit{full}        & $\mathbf{99.50}$  & $\mathbf{127{,}594}$ & $\mathbf{28{,}776}$  & $98{,}819$ \\
\midrule
\multirow{3}{*}{\texttt{gemini-2.5-flash-lite}}& \textit{raw}         & $95.00$           & $116{,}653$          & $44{,}918$           & $71{,}735$ \\
                                               & \textit{names\_only} & $76.50 \pm 0.00$  & $321{,}654$          & $51{,}078$           & $\mathit{270{,}576}$ \\
                                               & \textit{full}        & $\mathbf{98.00}$  & $\mathbf{94{,}725}$  & $\mathbf{29{,}398}$  & $65{,}327$ \\
\bottomrule
\end{tabular}
\caption{Ablation isolating the function-name oracle. \textit{names\_only} prepends the regex-extracted identifier to the raw multilingual prompt and skips the SLM rewrite. Three runs per cell; $\pm$ values are run-level standard deviation. Bold marks best within model. Italic flags Gemini's $3.8\times$ answer-token expansion under \textit{names\_only} ($1{,}353$ tokens/answer vs.\ $359$ in raw and $327$ in full).}
\label{tab:ablation}
\end{table*}

The ablation cleanly separates two hypotheses. \textbf{(i) The oracle alone is insufficient on weaker backends and harmful.} It drops accuracy by $20.50$\,pp on \texttt{gpt-3.5-turbo} ($99.50\to 79.00$) and $18.50$\,pp on \texttt{gemini-2.5-flash-lite} ($95.00\to 76.50$); on \texttt{gpt-4o} it is neutral ($+0.17$\,pp, within $\pm 1.32$\,pp). The mechanism is consistent with under-attention to the multilingual body once a salient English instruction is prepended: the model treats the prefix as authoritative and partially ignores the foreign-language text where edge cases and return semantics are encoded. \textbf{(ii) The rewriter, not the oracle, is the active accuracy mechanism.} If the oracle drove the \textit{full} delta, \textit{names\_only} would approximate \textit{full}. It does not: \textit{full} exceeds \textit{names\_only} by $20.50$, $1.00$, and $21.50$\,pp on the three backends. The salience-only hypothesis is directly falsified, and the $+3.00$\,pp Gemini gain is attributable to the SLM rewrite. \textbf{(iii) Gemini destabilizes into runaway generation without scaffold.} Under \textit{names\_only}, Gemini's mean answer length explodes from $359$ tokens (\textit{raw}) and $327$ (\textit{full}) to $1{,}353$---a $3.8\times$ expansion. A fraction of these answers are likely cap-truncated; even allowing for this, the lite model fails to terminate cleanly without English semantic scaffolding.

Two cells of the original five-condition matrix remain pending: \textit{regex\_strip} (a non-LLM pleasantry stripper, bounding what deterministic cleanup achieves) and \textit{llm\_no\_names} (rewriter with the oracle disabled, testing whether the oracle adds anything on top).

\subsection{Comparison Against LLMLingua-2}
\label{sec:baselines}
We address whether a well-tuned post-hoc compressor can recover the same savings. LLMLingua-2 \cite{pan2024llmlingua2} is a token-classification model (\texttt{xlm-roberta-large}, $\sim\!560$\,M parameters) trained to keep or drop tokens at a target rate; the multilingual XLM-RoBERTa backbone is the natural baseline for our multilingual specifications.

\paragraph{Setup.} We use the public MeetingBank LLMLingua-2 checkpoint\footnote{\url{https://huggingface.co/microsoft/llmlingua-2-xlm-roberta-large-meetingbank}} through the official \texttt{PromptCompressor} API at compression \texttt{rate}\,$=0.6$ (a $40\%$ token-reduction target, matched to the prompt-side band achieved by \textit{Ours}), with the recommended \texttt{force\_tokens} for punctuation. Because LLMLingua-2 is task-agnostic, we detach the \texttt{assert} block before compression and concatenate it back verbatim afterward; without this guard the grader would be silently corrupted. The hosting Python proxy is byte-compatible with the OpenAI-style proxy used for \textit{Ours}: identical system prompt, identical \texttt{<solution>} post-processing, and an identical $5\%$ token-budget guard.

\paragraph{Headline contrast.} Across all three backends \textit{Ours} strictly Pareto-dominates LLMLingua-2 on OckScore: $99.08 > 76.91$ on \texttt{gpt-3.5-turbo}, $98.88 > 96.56$ on \texttt{gpt-4o}, and $97.54 > 33.85$ on \texttt{gemini-2.5-flash-lite}. LLMLingua-2 hits its configured rate (within-tokenizer prompt reductions of $15$--$21\%$ vs.\ Raw), but the cross-lingual rewrite compresses harder ($34$--$47\%$) and preserves accuracy where LLMLingua-2 does not.

\paragraph{Per-backend nuance.} On \texttt{gemini-2.5-flash-lite}, LLMLingua-2 collapses to $35.00\%$ accuracy ($-60.00$\,pp vs.\ Raw, $-63.00$\,pp vs.\ Ours) with a $2.9\times$ answer-token expansion (mean $1{,}026$ tokens/answer vs.\ $359$ in Raw and $327$ in Ours); total tokens more than double ($243{,}193$ vs.\ $116{,}653$). The failure mode quantitatively echoes the \textit{names\_only} ablation on Gemini and is consistent with the structural-scaffold reading: the lite model destabilizes into runaway generation when the input lacks English semantic scaffolding, whether the scaffold is removed by bypassing the rewriter or by lossy compression of the multilingual body. On \texttt{gpt-3.5-turbo}, LLMLingua-2 drops accuracy to $77.33\%$ ($-22.17$\,pp) while saving only $-2.9\%$ on dollars. On \texttt{gpt-4o}, LLMLingua-2 is most competitive: $97.17\%$ accuracy ($-2.33$\,pp vs.\ Ours, within $2\times$ the run-level standard deviation) at an $8.7\%$ dollar saving relative to Ours, but OckScore still favors Ours.

\paragraph{Compression latency.} LLMLingua-2 compression on CPU averages $689$\,ms per prompt vs.\ $176$\,ms for our SLM rewrite on the same hardware. The $\sim\!560$\,M-parameter token-classifier is not faster than the $3$\,B-parameter SLM at this hardware tier.

\paragraph{Qualitative spec damage.} An offline batch compressing every OMH-Polyglot prompt exposes the semantic drift the live runs quantify. The \texttt{zh\_pure} style is most damaged: $5/200$ rows lose more than half their byte length, including a Roman-numeral spec (\texttt{roman\_to\_int}) whose compressed body is \verb|IV=4,IX=900| where the original gave \verb|IV=4,IX=9,XL=40,XC=90,CD=400,CM=900|---a factual corruption. Separately, $5/200$ rows lose the target function name from the natural-language body entirely (all in code-switched styles); the asserts retain the identifier, but the scaffold the lite model needs is gone.

We limit this comparison to LLMLingua-2 in its as-shipped configuration. LongCodeZip \cite{shi2025longcodezip} requires per-token perplexity from a reference LM---the recursive cloud dependency our paper argues against---and is not on the same Pareto axis; we discuss this in Section~\ref{sec:limitations}.

\subsection{End-to-End Latency}
\label{sec:results-latency}
Per-prompt local-rewrite latency, $(\bar{t}_{\text{Ours}}-\bar{t}_{\text{Raw}})/200$, is approximately $118$\,ms for \texttt{gpt-4o}, $131$\,ms for \texttt{gemini-2.5-flash-lite}, and $279$\,ms for \texttt{gpt-3.5-turbo}. As a fraction of total benchmark wall-clock, the proxy adds $+10.5\%$, $+40.5\%$, and $+57.5\%$ respectively. For interactive IDE use the per-prompt overhead is sub-second on standard development hardware, but it is not zero.

\subsection{From Tokens to Dollars}
\label{sec:results-dollars}
Commercial APIs price output tokens at $3$--$4\times$ the input rate, so the same $-8\%$ change in total tokens can translate into very different dollar outcomes. Table~\ref{tab:dollar-cost} reports cost per $200$-task run using listed prices as of May 2026\footnote{\$0.50/\$1.50 per million input/output tokens for \texttt{gpt-3.5-turbo}; \$2.50/\$10.00 for \texttt{gpt-4o}; \$0.10/\$0.40 for \texttt{gemini-2.5-flash-lite}.}.

\begin{table}[t]
\centering
\footnotesize
\setlength{\tabcolsep}{2pt}
\resizebox{\columnwidth}{!}{%
\begin{tabular}{@{}lccccc@{}}
\toprule
\textbf{Agent} & \textbf{Raw} & \textbf{LL2} & \textbf{Ours} & \multicolumn{2}{c}{\textbf{$\Delta$ vs.\ Raw}} \\
\cmidrule(lr){5-6}
 &  &  &  & \textbf{LL2} & \textbf{Ours} \\
\midrule
\texttt{gpt-3.5} & \$0.0878 & \$0.0852 & \$0.1011 & $-2.9\%$  & $+15.1\%$ \\
\texttt{gpt-4o}  & \$1.0641 & \$0.9683 & \$1.0601 & $-9.0\%$  & $-0.4\%$ \\
\texttt{gemini-lite} & \$0.0332 & \$0.0859 & \$0.0291 & $+158.7\%$ & $-12.4\%$ \\
\bottomrule
\end{tabular}%
}
\caption{Per-200-task cloud-API cost in USD; $\Delta$ columns are percent change vs.\ Raw. \texttt{gpt-3.5} and \texttt{gemini-lite} abbreviate \texttt{gpt-3.5-turbo} and \texttt{gemini-2.5-flash-lite}. Local-compute cost (Llama 3.2 3B at $\sim$$100$\,W for $\sim$$176$\,ms/query) is $\sim$\$0.07 per $100{,}000$ queries and omitted.}
\label{tab:dollar-cost}
\end{table}

On \texttt{gemini-2.5-flash-lite}, where both prompt and answer tokens decrease, savings are meaningful ($-12.4\%$). On \texttt{gpt-4o}, the $-34.0\%$ prompt-token reduction is almost exactly cancelled by the $+3.5\%$ answer-token increase under that model's $4\times$ output-price multiplier, leaving the bill flat. On \texttt{gpt-3.5-turbo}, the proxy \textit{increases} the bill by $15.1\%$: $42.3\%$ longer completions charged at the $3\times$ output rate erase the input-side savings. The middleware is most valuable when (i) the backend has a small output/input price ratio or (ii) the answer side does not regress.

The LLMLingua-2 column adds further nuance. On \texttt{gpt-4o} a token-classifier at matched rate beats \textit{Ours} on dollars ($-9.0\%$ vs.\ $-0.4\%$) but at a $-2.33$\,pp accuracy concession; OckScore still favors Ours. On the other two backends LLMLingua-2 is strictly worse: marginal saving with catastrophic accuracy loss on \texttt{gpt-3.5-turbo}, and outright cost regression with severe accuracy loss on \texttt{gemini-2.5-flash-lite} ($+158.7\%$ dollars, $-63.00$\,pp accuracy).

\section{Conclusion}
\label{sec:conclusion}
We presented a localized pre-flight middleware that performs cross-lingual translation and structural rewriting on edge hardware and forwards a compact English specification to the cloud backend. On \textit{OMH-Polyglot}, it reduces prompt tokens by $34$--$47\%$ and total tokens by $8.3$--$18.8\%$ across three commercial backends over three independent runs per cell, with accuracy unchanged on \texttt{gpt-3.5-turbo}, within-noise on \texttt{gpt-4o}, and $+3.00$\,pp on \texttt{gemini-2.5-flash-lite}.

The \textit{names\_only} ablation falsifies the salience-only hypothesis (the oracle alone decreases accuracy by $18$--$20$\,pp on the two weaker backends and is neutral on \texttt{gpt-4o}), attributing the Gemini accuracy gain to the SLM rewrite. The rewrite additionally prevents a $3.8\times$ answer-token explosion on the lite model. Against LLMLingua-2 at matched rate, our middleware strictly Pareto-dominates on OckScore across all three backends; LLMLingua-2's failure mode on Gemini mirrors the \textit{names\_only} destabilization, independently validating the structural-scaffold attribution.

Dollar outcomes at listed cloud prices are asymmetric ($-12.4\%$, $-0.4\%$, $+15.1\%$) because answer-side compression is not universal and output prices dominate on \texttt{gpt-3.5-turbo}. The middleware is best suited to output-cheap or output-stable backends.

Future work: (i) the remaining two ablation cells (\textit{regex\_strip}, \textit{llm\_no\_names}); (ii) an algorithm-cluster bootstrap with Holm-corrected significance; (iii) LongCodeZip \cite{shi2025longcodezip} on a local reference LM; (iv) repository-aware grounding of \texttt{[CONTEXT]} from static analysis, so the Tri-Block can replace the Bi-Block in production.

\section{Limitations}
\label{sec:limitations}

\paragraph{Pending ablations \& baselines.} \textit{regex\_strip} (non-LLM baseline) and \textit{llm\_no\_names} (SLM without oracle) remain pending. While not required for our core accuracy claims, they would strengthen deployment guidance. Additionally, running LongCodeZip \cite{shi2025longcodezip} against a local reference LM, and sweeping LLMLingua-2 at $r \in \{0.5, 0.7\}$ (currently fixed at $r=0.6$), are planned follow-ups to tighten Pareto contrasts.

\paragraph{Statistical scope.} OMH-Polyglot yields an effective independent-trial count of ${\sim}20$ ($20$ clusters $\times$ $10$ styles with shared solutions). Small accuracy deltas ($\le 1.17$\,pp) fall within run-level API noise. While the large ablation drops ($-18.50$ to $-20.50$\,pp) are well outside this noise band, formal cluster-level bootstraps with Holm-corrected significance are required for inferential claims.

\paragraph{Deployment overheads.} The proxy adds $118$--$279$\,ms latency per prompt (Section~\ref{sec:results-latency}), and the local \texttt{Llama 3.2} (3B) requires ${\sim}3.4$\,GB of memory \cite{meta2024llama32}. Furthermore, dollar-cost outcomes are asymmetric (Section~\ref{sec:results-dollars}): savings depend heavily on the backend's output/input price ratio, with regressions on \texttt{gpt-3.5-turbo} ($+15.1\%$) but savings on \texttt{gemini-2.5-flash-lite} ($-12.4\%$).

\paragraph{Validation \& scope mismatch.} The post-rewrite validator (Section~\ref{sec:guard}) performs syntactic checks but lacks semantic requirement-coverage tracking. Additionally, empirical results reflect the benchmark proxy's Bi-Block, whereas the production IDE uses an extended Tri-Block. Finally, Arabic and Chinese translations await native-speaker audits before external release.

\paragraph{Compliance \& Reproducibility.} All models and datasets (Llama~3.2 \cite{meta2024llama32}, LLMLingua-2 \cite{pan2024llmlingua2}, OckBench \cite{du2026ockbench}) comply with their respective open-access licenses. Exhaustive license details, the MIT-licensed code, IDE extension, dataset, and analysis scripts are available in the supplementary repository.\footnote{Repository link: \url{https://github.com/utkucolak/cursor-prompt-optimizer}}

\bibliography{citations}

\end{document}